\pdfoutput=1

\documentclass[11pt]{article}

\usepackage{acl}

\usepackage{times}
\usepackage{latexsym}
\usepackage{multirow}
\usepackage{booktabs}
\usepackage{subcaption}
\usepackage[T1]{fontenc}

\usepackage[utf8]{inputenc}

\usepackage{microtype}
\usepackage{graphicx}
\title{Automatic Anonymization of Swiss Federal Supreme Court Rulings}

\author{
Joel Niklaus$^{1,2,3\;}\thanks{\hspace{2mm}Equal contribution.}$  \And
Robin Mamié$^{4\;*}$  \AND
Matthias Stürmer$^{1,2}$ \And
Daniel Brunner$^{4}$ \And
Marcel Gygli$^{2}$\AND
\\
$^1$University of Bern\quad
$^2$Bern University of Applied Sciences\\
$^3$Stanford University\quad
$^4$Swiss Federal Supreme Court\\
}

\begin{document}
\maketitle
\begin{abstract}
Releasing court decisions to the public relies on proper anonymization to protect all involved parties, where necessary.
The Swiss Federal Supreme Court relies on an existing system that combines different traditional computational methods with human experts.
In this work, we enhance the existing anonymization software using a large dataset annotated with entities to be anonymized.
We compared BERT-based models with models pre-trained on in-domain data.
Our results show that using in-domain data to pre-train the models further improves the F1-score by more than 5\% compared to existing models.
Our work demonstrates that combining existing anonymization methods, such as regular expressions, with machine learning can further reduce manual labor and enhance automatic suggestions.
\end{abstract}




\begin{figure*}[ht]
    \centering
    \includegraphics{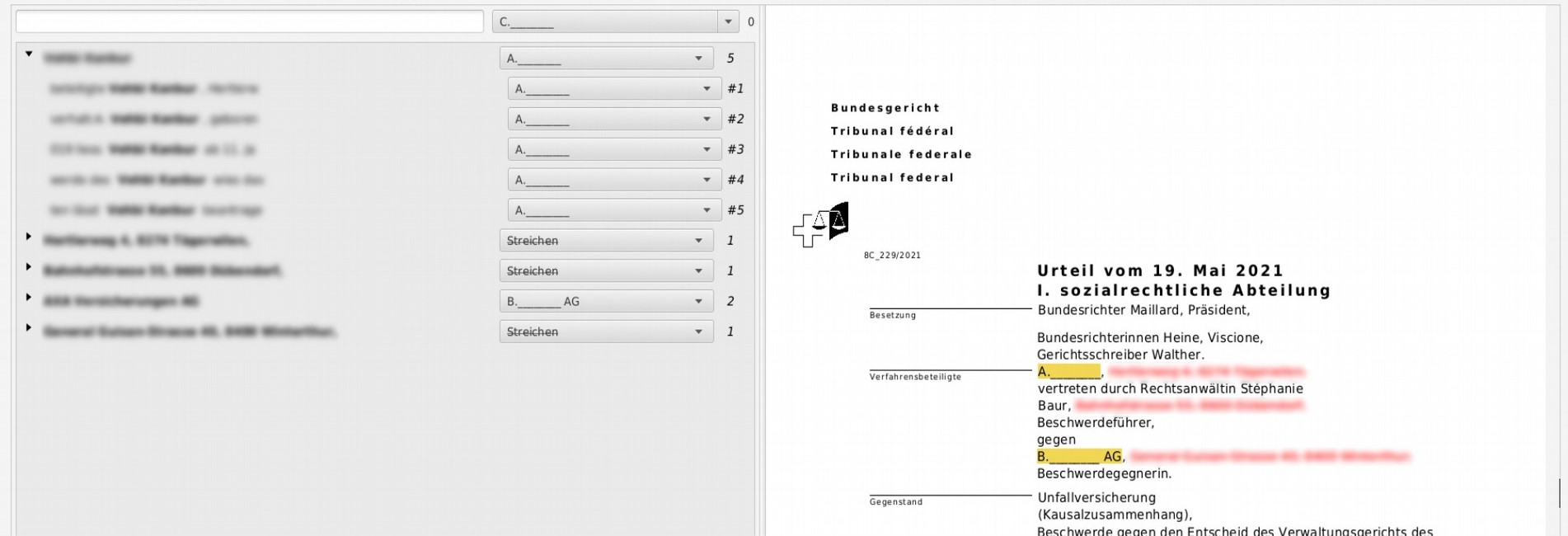}
    \caption{Main window of \textit{Anom2}. Anonymizations are configured on the left, and the anonymized court ruling appears on the right. The system highlights completed anonymizations in gold and the current setting in yellow.}
    \label{fig:anom_main}
\end{figure*}

\section{Introduction}

The Swiss Federal Supreme Court (SFSC) is the highest judicial authority in Switzerland. It is the final arbiter in legal disputes and ensures the uniform application of federal law throughout the country. It consists of several divisions specialized in different areas of law, including civil, criminal, administrative, and social security matters \cite{glaser_anonymization_2021}.
In a year, the SFSC roughly handles 7K cases and publishes its rulings. In this process, personal information must be anonymized from the rulings in order to protect involved parties. In the traditional setting, court rulings are anonymized by skilled experts. This task is highly complex, as the removal/anonymization of a word is dependent on the context it is written in. For example, \textit{Zuerich} needs to be removed if it is part of the name of the legal entity "Zurich Insurance Group", but not if it is a reference to the city.
At the SFSC, experts are already supported in their work through an application called \textit{Anom2} (see Figure \ref{fig:anom_main}).
\textit{Anom2} provides access to various methods and algorithms for finding and replacing text entities (e.g., with regular expressions).
The aim of this work is to enhance the capabilities of \textit{Anom2} with Machine Learning capabilities that provide the user with more suggestions that need to be anonymized. Our results show that this approach allows users to find more elements that require anonymization.

\section{Related Work}
For identifying elements that might require anonymization, a process called Named Entity Recognition (NER) is employed.
Traditionally, NER recognizes and categorizes text parts according to a set of semantic categories like \textit{Location (LOC)}, \textit{Organization (ORG)}, or \textit{Person (PER)}~\cite{benikova_nosta-d_2014}.
As these classes are not enough for the anonymization of court cases~\cite{leitner_dataset_2020} suggested expanding this list to seven coarse and 19 fine-grained classes, including entities such as \textit{Judge (RR)}, or \textit{Lawyer (AN)}. Using this dataset,~\citet{darji_german_2023} fine-tuned GermanBERT~\cite{chan_germans_2020}, clearly outperforming a BiLSTM-CRF+ model. Similar approaches have been applied and tested in other languages, such as Romanian~\cite{pais_named_2021}, Greek~\cite{angelidis_named_2018}, Portuguese~\cite{luz_de_araujo_lener-br_2018}, and multilingually~\cite{mapa, niklaus2023lextreme}.

Domain-specific pretraining has flourished in the legal domain recently. \citet{chalkidis_legal-bert_2020} pretrained LegalBERT on EU and UK legislation, ECHR and US cases, and US contracts. \citet{zheng_when_2021} pretrained CaseHoldBERT on US case law, while \citet{henderson_pile_2022} trained PoL-BERT on the 256 GB Pile of Law corpus. \citet{niklaus_budgetlongformer_2022} pretrained Longformer \cite{beltagy2020longformer} models using the Replaced Token Detection (RTD) task on the Pile of Law. \citet{hua_legalrelectra_2022} used RTD to pretrain Reformer \cite{kitaev_reformer_2020} models on 6 GB of US case law. Finally, \citet{niklaus_multilegalpile_2023} released a large multilingual legal corpus and trained various legal models. We continue pretraining the German, French, and Italian models for 800K and 300K steps more for base and large models, respectively. \citet{rasiah_scale_2023} pretrain models on Swiss legal data, termed Legal-Swiss-RoBERTa. 

Document anonymization has a long tradition in the medical domain, where personal data need to be removed from documents. 
Initially, this task was handled using methods such as semantic lexicons~\cite{ruch_medical_2000} or regular expressions to replace text occurrences.
Recently, this has been expanded to include BERT-style models as well~\cite{mao_hadoken_2019}.
In the legal domain, \citet{glaser_anonymization_2021} worked on 1400 anonymized German rulings. Using already anonymized rulings, they trained different Recurrent Neural Networks (RNN) using BERT embeddings. Using this approach, they achieved a maximum of 68.9\% precision and 79.1\% recall rates.
\citet{garat_automatic_2022} performed similar work on 80K documents from Uruguayan courts.
Our work specifically tackles court decisions by the SFSC. We compare the generic cased mBERT model \cite{devlin-etal-2019-bert} with models pre-trained on in-domain data (such as Legal-Swiss-RoBERTa-base \cite{rasiah_scale_2023}).
We also investigate monolingual model performance in the three languages of the SFSC rulings: German, French, and Italian.

Much prior work used SFSC cases as data for their research because of wide availability in three languages, giving good coverage of the most important Swiss case law. \citet{niklaus_swiss-judgment-prediction_2021, niklaus_empirical_2022} introduced and studied judgment prediction on SFSC rulings. \citet{brugger_multilegalsbd_2023} investigated and improved multilingual sentence boundary detection in the legal domain using SFSC decisions. \citet{christen_resolving_2023} studied negation scope resolution and \citet{nyffenegger_anonymity_2023} investigated how easily LLMs can re-identify persons occurring in anonymized SFSC decisions. \citet{rasiah_scale_2023} created a large benchmark of ten text classification tasks, two text generation tasks, an information retrieval, and a citation extraction task.

\begin{figure*}[ht]
    \centering
    \begin{subfigure}[b]{0.45\textwidth}
        \includegraphics[width=\textwidth]{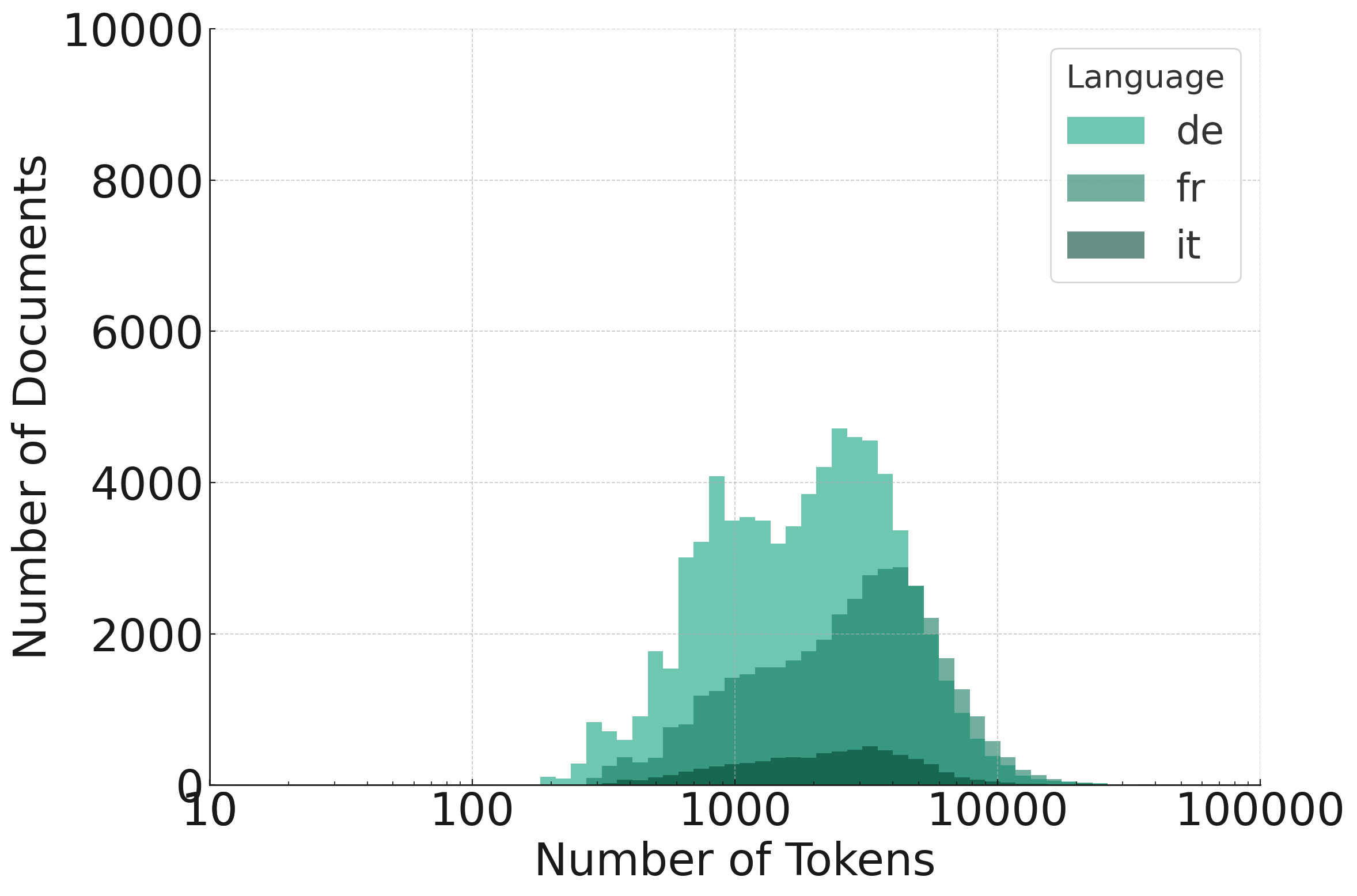}
        \caption{Token Distribution}
        \label{fig:tokens}
    \end{subfigure}
    \hfill
    \begin{subfigure}[b]{0.45\textwidth}
        \includegraphics[width=\textwidth]{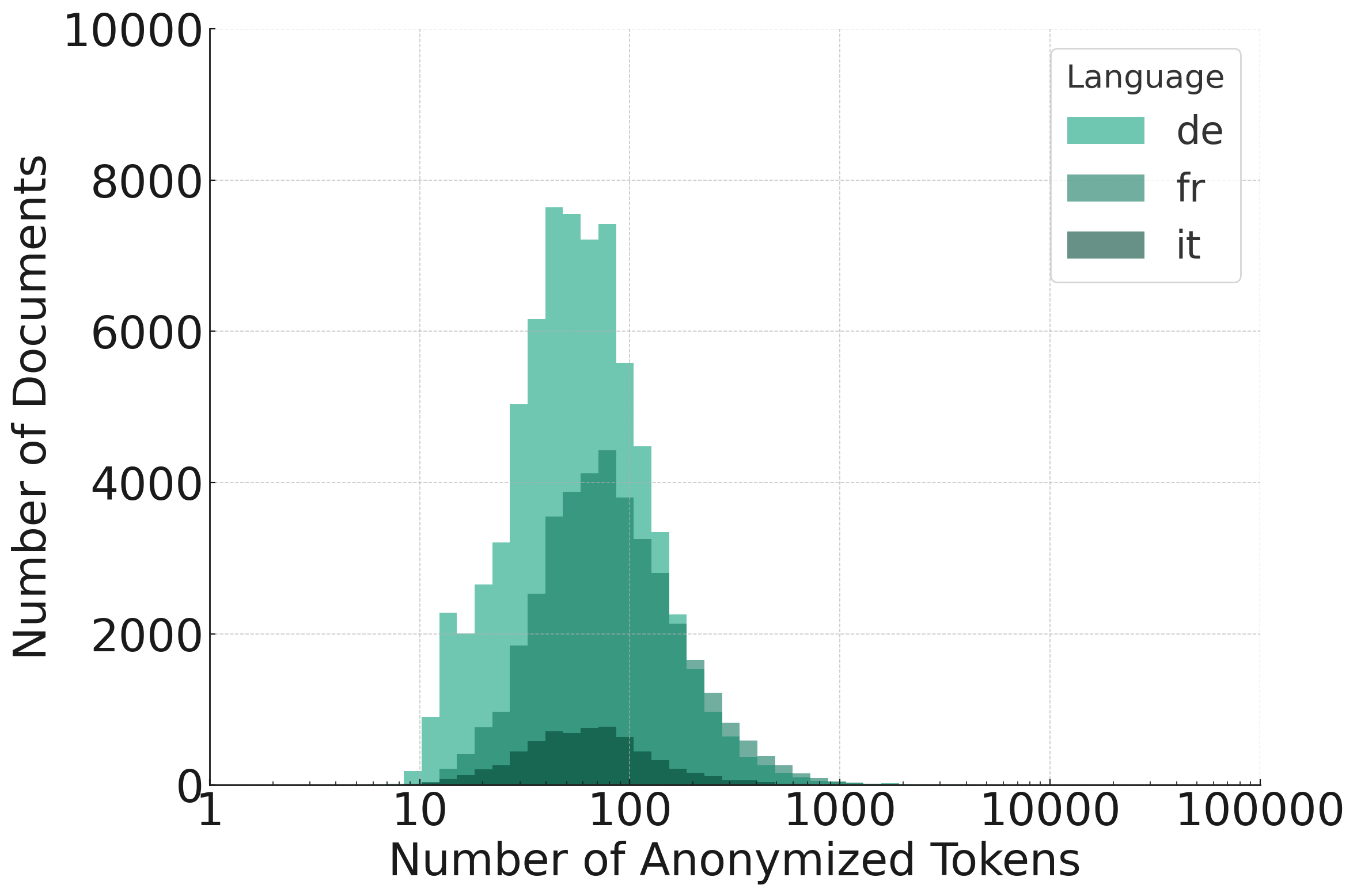}
        \caption{Anonymized Token Distribution}
        \label{fig:anonymized_tokens}
    \end{subfigure}
    
    \vskip\baselineskip
    
    \begin{subfigure}[b]{0.45\textwidth}
        \includegraphics[width=\textwidth]{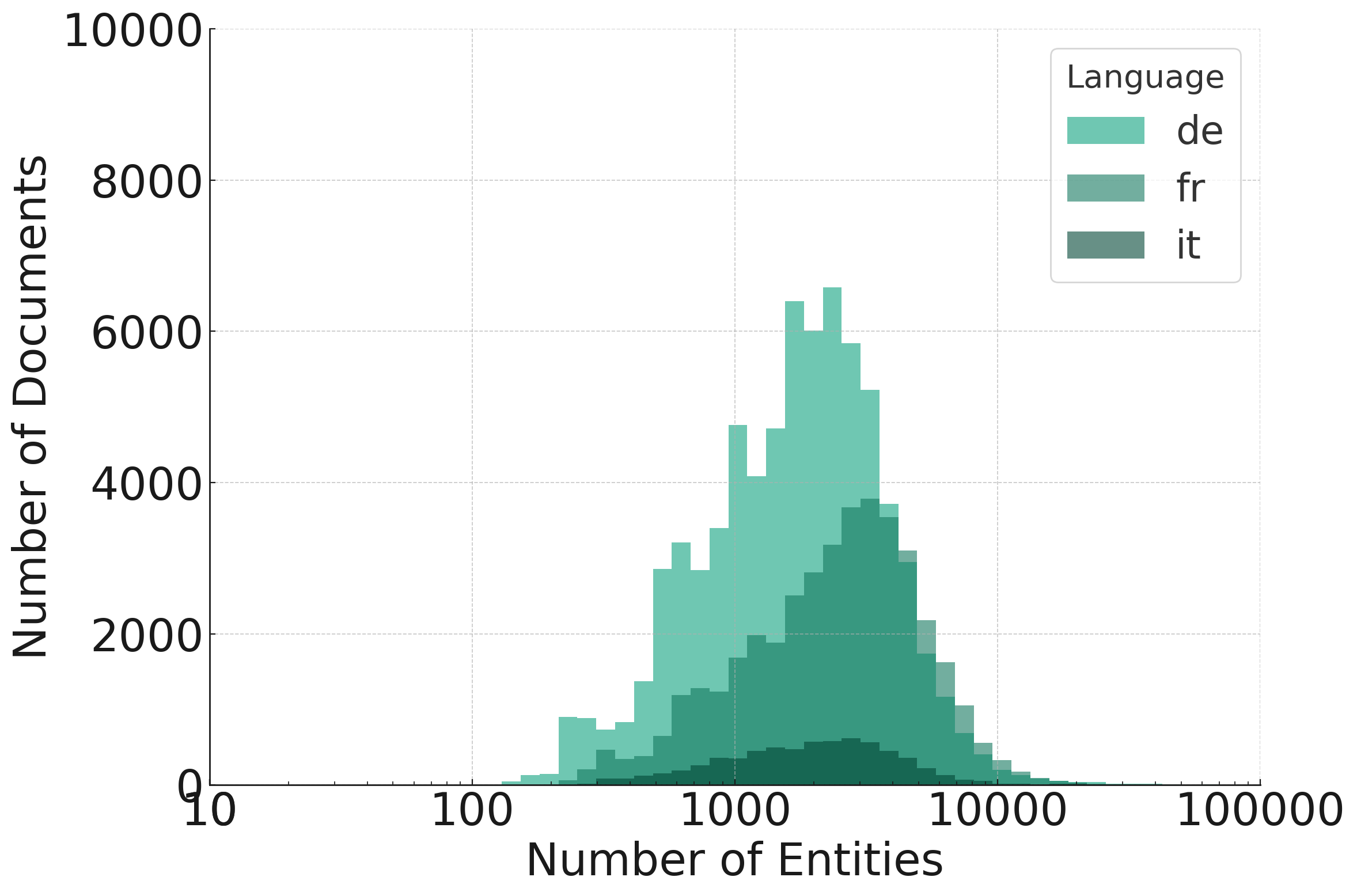}
        \caption{Entity Distribution}
        \label{fig:entities}
    \end{subfigure}
    \hfill
    \begin{subfigure}[b]{0.45\textwidth}
        \includegraphics[width=\textwidth]{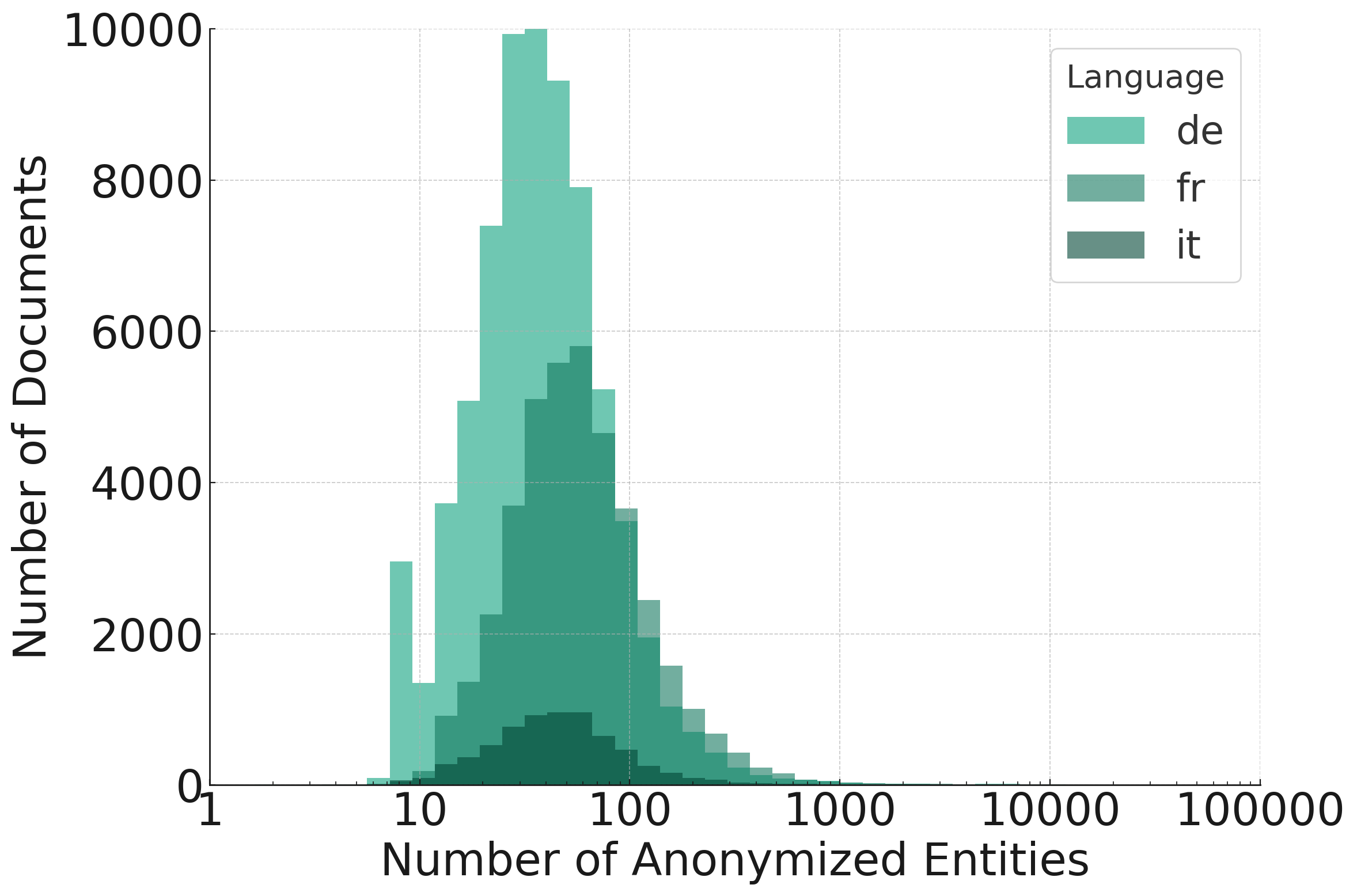}
        \caption{Anonymized Entity Distribution}
        \label{fig:anonymized_entities}
    \end{subfigure}
    \caption{Histograms illustrating the distribution of (anonymized) tokens and entities across the three languages.}
    \label{fig:all_measures}
\end{figure*}

\section{Dataset}
We used 119156 rulings (77262 German, 40099 French, 6795 Italian) Supreme court decisions and split them into sentences using Spacy \cite{honnibal_spacy_2020}. We prepared the decisions for NER based on the manual labels from the paralegals who performed manual anonymizations.
The histograms in Figure \ref{fig:all_measures} illustrate the distribution of four key measures, namely, Number of Tokens, Anonymized Tokens, Entities, and Anonymized Entities, in three languages: German (de), French (fr), and Italian (it). Different color schemes for each language enhance the visual interpretability of the plots. Measures concerning tokens and entities exhibit a long-tailed distribution, signifying a concentration of instances at the lower end of the value spectrum. Specifically, the distribution of Number of Tokens and Number of Entities is examined within a 10 to 100,000 range, capturing their broad spread. In contrast, anonymized tokens and entities are evaluated within a 1 to 10,000 range, reflecting their constrained distribution.


\begin{table*}[ht]
\centering
\caption{Evaluation Results. Best results per setup are in \textbf{bold}.}
\label{tab:evaluation_results}
\resizebox{\textwidth}{!}{
\begin{tabular}{lccc|ccc}
\toprule
\multirow{2}{*}{Model} & \multicolumn{3}{c|}{Normal} & \multicolumn{3}{c}{Uniformizing} \\
\cline{2-7}
& Precision & Recall & F1-Score & Precision & Recall & F1-Score \\
\midrule
\bf Multilingual Models\\
bert-base-multilingual-cased & 90.72 & 83.76 & 87.10 & 85.85 & \bf 94.95 & \bf 90.17 \\
Legal-XLM-RoBERTa-base & \bf 94.84 & 81.98 & 87.94 & \bf 89.93 & 86.85 & 88.36 \\
Legal-Swiss-RoBERTa-base & 92.26 & \bf 92.57 & \bf 92.42 & 83.13 & 94.85 & 88.60 \\
\midrule
\midrule
\bf Monolingual Models\\
bert-base-german-cased & 95.14 & 80.00 & 86.92 & \bf 91.49 & \bf 85.86 & \bf 88.58 \\
Legal-German-RoBERTa-base & \bf 95.40 & \bf 80.09 & \bf 87.07 & 89.20 & 84.97 & 87.03 \\
\midrule
dbmdz/bert-base-french-europeana-cased & \bf 95.86 & 81.84 & 88.30 & \bf 88.92 & 89.14 & \bf 89.03 \\
Legal-French-RoBERTa-base & 95.45 & \bf 83.48 & \bf 89.06 & 88.77 & \bf 89.17 & 88.97 \\
\midrule
dbmdz/bert-base-italian-cased & 93.49 & 80.21 & 86.35 & 76.71 & 83.85 & 80.12 \\
Legal-Italian-RoBERTa-base & \bf 94.16 & \bf 80.59 & \bf 86.85 & \bf 84.03 & \bf 84.06 & \bf 84.05 \\
\bottomrule
\end{tabular}
}
\end{table*}

\section{Legal Pretraining}

To improve the SFSC anonymization system, we pretrained legal-specific models on diverse legal text in German, French, and Italian.

\noindent (a) We warm-start (initialize) our models from the original XLM-R checkpoints (base or large) of \citet{conneau_cross-lingual_2019}. Model recycling is a standard process followed by many~\cite{wei-etal-2022, instructgpt} to benefit from starting from an available ``well-trained'' PLM, rather from scratch (random). XLM-R was trained on 2.5~TB of cleaned CommonCrawl data in 100 languages. 

\noindent (b) We train a new tokenizer of 32K BPEs on the training subsets to better cover legal language. However, we reuse the original XLM-R embeddings for all lexically overlapping tokens \cite{pfeiffer-etal-2021-unks}, i.e., we warm-start word embeddings for tokens that already exist in the original XLM-R vocabulary, and use random ones for the rest.

\noindent (c) We continue pretraining our monolingual models on our pretraining corpus with batches of 512 samples for an additional 1M/500K steps for the base/large model. We do initial warm-up steps for the first 5\% of the total training steps with a linearly increasing learning rate up to $1e\!-\!4$, and then follow a cosine decay scheduling, following recent trends. For half of the warm-up phase (2.5\%), the Transformer encoder is frozen, and only the embeddings, shared between input and output (MLM), are updated. We also use an increased 20/30\% masking rate for base/large models respectively, where also 100\% of the predictions are based on masked tokens, compared to \citet{devlin-etal-2019-bert}\footnote{\citet{devlin-etal-2019-bert} -- and much follow-up work -- used a 15\% masking ratio, and a recipe of 80/10/10\% of predictions made across masked/randomly-replaced/original tokens.}, based on the findings of~\citet{wettig2022should}.

\noindent (d) We consider mixed cased models, i.e., both upper- and lowercase letters covered, similar to recently developed large PLMs~\cite{conneau_cross-lingual_2019, raffel_exploring_2020, brown-etal-gpt3}.

\noindent (e) This leaves us with two models for each language (base and large). Additionally, we consider the multilingual legal models pretrained by \citet{niklaus_multilegalpile_2023} and the Swiss legal models pretrained by \citet{rasiah_scale_2023}.

\section{Anonymization System}
The SFSC employs an anonymization system, \textit{Anom2}, to assist paralegals in anonymizing rulings for public access. The main UI is shown in Figure \ref{fig:anom_main}. Upon loading a ruling, the application auto-identifies terms requiring anonymization and lists them on the left, along with replacement text. The search function allows direct term marking for anonymization. 
\textit{Anom2} uses different algorithms for the search for text that needs to be anonymized:\\
\textbf{Conventional} is based on a statistical analysis of the loaded ruling. Using \textit{polyglot}\footnote{See: \url{https://polyglot.readthedocs.io}} an initial set of named entities is detected. Using the specific knowledge of the format, the rubrum is dynamically detected, allowing for the labelling of important names and addresses.\\
\textbf{BERT} performs the recognition of entities to be anonymized using a BERT~\cite{devlin-etal-2019-bert} model fine-tuned for NER. Entity recognition is performed on the sentence level, as the rulings are often too long for the model. 
This approach could lead to inconsistencies in recognition, as a term identified in one sentence might not be identified in another. This is solved in post-processing, where any identified term is automatically anonymized in the whole document.\\
\textbf{Legal-Swiss-RoBERTa-base} works analogously to the BERT method, but uses a fine-tuned \textit{Legal-Swiss-RoBERTa-base}~\cite{rasiah_scale_2023} model. 


\section{Experimental Setup}

We used the following hyperparameters for all evaluated models: batch size of 64, learning rate of 5e-5, and weight decay of 0.01. We employed the seqeval metric for evaluation. We set the maximum sequence length to 192 tokens, which we determined to be the optimal trade-off between average sentence size and training time for computational efficiency. We used early stopping based on the F1-score of the validation set, which constitutes 10\% of the entire dataset, following an 80-10-10 split for the training, validation, and test sets, respectively. Training ceases once the F1-score on the validation set starts to decline. Due to resource constraints (we only had two Tesla T4 GPUs) we could only run one random seed per model.

We define and configure two special parameters:
1) \emph{TruncationStrideRatio}: We set this parameter to 0.5. When a sentence exceeds 192 tokens, we truncate it using a specific overlap strategy. The overlap consists of half of the previous snippet and half of the next snippet.\\
2) \emph{NonAnonymizedSentencesRatioToAnonymizedSentences}: We set the ratio at 1.5, including only 150\% of sentences without anonymization examples compared to those with examples. This minimizes data redundancy and maximizes utility.

\section{Results}
Table \ref{tab:evaluation_results} presents a comprehensive evaluation of various BERT and RoBERTa-based models on two different conditions: Normal and Uniformizing. 
For the Normal condition, in the multilingual setting, Legal-XLM-RoBERTa-base exhibits the highest Precision at 94.84\%, while Legal-Swiss-RoBERTa-base demonstrates superior Recall and F1-Score values, achieving 92.57\% and 92.42\% respectively. 
With Uniformizing, we describe the process of forcing the model to replace all occurrences of a detected term across the whole document.
This approach leads to better Recall, but reduces Precision.
In the Uniformized case, again Legal-XLM-RoBERTa-base shows highest Precision, while mBERT achieves highest Recall and F1-Score. 
The improved Recall and F1-Score in the Normal condition show that pre-training on legal data can improve the performance of models. 
We observe similar behavior for the monolingual models. 
All models pre-trained on legal data achieve a higher F1-Score than generic monolingual models. 

\section{Discussion}
We pretrained models on Swiss legal data and performed a detailed comparison of legal and generic models, both multilingually and monolingually in the ruling anonymization task. Our experiments indicate that pretraining on legal data improves the performance of models significantly compared to generic multi- or monolingual models. 

To reduce errors in sentence splitting, we suggest future work to use legal specific sentence splitters \cite{brugger_multilegalsbd_2023}. Due to computational constraints we only experimented with base size encoder models. Future work may expand this by also testing larger models.



\section*{Acknowledgements}
We greatly appreciate Google's generous support of TPUs v3-8 machines for pretraining the models. This work has been partially supported by the Swiss National Research Programme “Digital Transformation” (NRP-77) grant number 187477.

\bibliography{references_marcel,references,custom}

\begin{thebibliography}{36}
\expandafter\ifx\csname natexlab\endcsname\relax\def\natexlab#1{#1}\fi

\bibitem[{Angelidis et~al.(2018)Angelidis, Chalkidis, and
  Koubarakis}]{angelidis_named_2018}
I.~Angelidis, Ilias Chalkidis, and M.~Koubarakis. 2018.
\newblock Named {Entity} {Recognition}, {Linking} and {Generation} for {Greek}
  {Legislation}.
\newblock In \emph{{JURIX}}.

\bibitem[{Beltagy et~al.(2020)Beltagy, Peters, and
  Cohan}]{beltagy2020longformer}
Iz~Beltagy, Matthew~E Peters, and Arman Cohan. 2020.
\newblock Longformer: The long-document transformer.
\newblock \emph{arXiv preprint arXiv:2004.05150}.

\bibitem[{Benikova et~al.(2014)Benikova, Biemann, and
  Reznicek}]{benikova_nosta-d_2014}
Darina Benikova, Chris Biemann, and Marc Reznicek. 2014.
\newblock \href
  {http://www.lrec-conf.org/proceedings/lrec2014/pdf/276_Paper.pdf}
  {{NoSta}-{D} {Named} {Entity} {Annotation} for {German}: {Guidelines} and
  {Dataset}}.
\newblock In \emph{Proceedings of the {Ninth} {International} {Conference} on
  {Language} {Resources} and {Evaluation} ({LREC}'14)}, pages 2524--2531,
  Reykjavik, Iceland. European Language Resources Association (ELRA).

\bibitem[{Brown et~al.(2020)Brown, Mann, Ryder, Subbiah, Kaplan, Dhariwal,
  Neelakantan, Shyam, Sastry, Askell, Agarwal, Herbert-Voss, Krueger, Henighan,
  Child, Ramesh, Ziegler, Wu, Winter, Hesse, Chen, Sigler, Litwin, Gray, Chess,
  Clark, Berner, McCandlish, Radford, Sutskever, and Amodei}]{brown-etal-gpt3}
Tom Brown, Benjamin Mann, Nick Ryder, Melanie Subbiah, Jared~D Kaplan, Prafulla
  Dhariwal, Arvind Neelakantan, Pranav Shyam, Girish Sastry, Amanda Askell,
  Sandhini Agarwal, Ariel Herbert-Voss, Gretchen Krueger, Tom Henighan, Rewon
  Child, Aditya Ramesh, Daniel Ziegler, Jeffrey Wu, Clemens Winter, Chris
  Hesse, Mark Chen, Eric Sigler, Mateusz Litwin, Scott Gray, Benjamin Chess,
  Jack Clark, Christopher Berner, Sam McCandlish, Alec Radford, Ilya Sutskever,
  and Dario Amodei. 2020.
\newblock \href
  {https://proceedings.neurips.cc/paper/2020/file/1457c0d6bfcb4967418bfb8ac142f64a-Paper.pdf}
  {Language models are few-shot learners}.
\newblock In \emph{Advances in Neural Information Processing Systems},
  volume~33, pages 1877--1901. Curran Associates, Inc.

\bibitem[{Brugger et~al.(2023)Brugger, Stürmer, and
  Niklaus}]{brugger_multilegalsbd_2023}
Tobias Brugger, Matthias Stürmer, and Joel Niklaus. 2023.
\newblock \href {https://doi.org/10.48550/arXiv.2305.01211} {{MultiLegalSBD}:
  {A} {Multilingual} {Legal} {Sentence} {Boundary} {Detection} {Dataset}}.
\newblock ArXiv:2305.01211 [cs].

\bibitem[{Chalkidis et~al.(2020)Chalkidis, Fergadiotis, Malakasiotis, Aletras,
  and Androutsopoulos}]{chalkidis_legal-bert_2020}
Ilias Chalkidis, Manos Fergadiotis, Prodromos Malakasiotis, Nikolaos Aletras,
  and Ion Androutsopoulos. 2020.
\newblock {LEGAL}-{BERT}: {The} {Muppets} straight out of {Law} {School}.
\newblock In \emph{Findings of the {Association} for {Computational}
  {Linguistics}: {EMNLP} 2020}, pages 2898--2904.

\bibitem[{Chan et~al.(2020)Chan, Schweter, and Möller}]{chan_germans_2020}
Branden Chan, Stefan Schweter, and Timo Möller. 2020.
\newblock \href {http://arxiv.org/abs/2010.10906} {German's {Next} {Language}
  {Model}}.
\newblock \emph{arXiv:2010.10906 [cs]}.
\newblock ArXiv: 2010.10906.

\bibitem[{Christen et~al.(2023)Christen, Shaitarova, Stürmer, and
  Niklaus}]{christen_resolving_2023}
Ramona Christen, Anastassia Shaitarova, Matthias Stürmer, and Joel Niklaus.
  2023.
\newblock \href {https://arxiv.org/abs/2309.08695v1} {Resolving {Legalese}: {A}
  {Multilingual} {Exploration} of {Negation} {Scope} {Resolution} in {Legal}
  {Documents}}.

\bibitem[{Conneau and Lample(2019)}]{conneau_cross-lingual_2019}
Alexis Conneau and Guillaume Lample. 2019.
\newblock \href
  {https://proceedings.neurips.cc/paper/2019/hash/c04c19c2c2474dbf5f7ac4372c5b9af1-Abstract.html}
  {Cross-lingual {Language} {Model} {Pretraining}}.
\newblock In \emph{Advances in {Neural} {Information} {Processing} {Systems}},
  volume~32. Curran Associates, Inc.

\bibitem[{Darji et~al.(2023)Darji, Mitrović, and
  Granitzer}]{darji_german_2023}
Harshil Darji, Jelena Mitrović, and Michael Granitzer. 2023.
\newblock \href {https://doi.org/10.5220/0011749400003393} {German {BERT}
  {Model} for {Legal} {Named} {Entity} {Recognition}:}.
\newblock In \emph{Proceedings of the 15th {International} {Conference} on
  {Agents} and {Artificial} {Intelligence}}, pages 723--728, Lisbon, Portugal.
  SCITEPRESS - Science and Technology Publications.

\bibitem[{de~Gibert et~al.(2022)de~Gibert, Garc{\'\i}a-Pablos, Cuadros, and
  Melero}]{mapa}
Ona de~Gibert, A~Garc{\'\i}a-Pablos, Montse Cuadros, and Maite Melero. 2022.
\newblock Spanish datasets for sensitive entity detection in the legal domain.
\newblock In \emph{Proceedings of the Thirteenth International Conference on
  Language Resources and Evaluation (LREC’22), Marseille, France, june.
  European Language Resource Association (ELRA)}.
\newblock Dataset URL: https://tinyurl.com/mv65cp66.

\bibitem[{Devlin et~al.(2019)Devlin, Chang, Lee, and
  Toutanova}]{devlin-etal-2019-bert}
Jacob Devlin, Ming-Wei Chang, Kenton Lee, and Kristina Toutanova. 2019.
\newblock \href {https://doi.org/10.18653/v1/N19-1423} {{BERT}: Pre-training of
  deep bidirectional transformers for language understanding}.
\newblock In \emph{Proceedings of the 2019 Conference of the North {A}merican
  Chapter of the Association for Computational Linguistics: Human Language
  Technologies, Volume 1 (Long and Short Papers)}, pages 4171--4186,
  Minneapolis, Minnesota. Association for Computational Linguistics.

\bibitem[{Garat and Wonsever(2022)}]{garat_automatic_2022}
Diego Garat and Dina Wonsever. 2022.
\newblock \href {https://doi.org/10.3390/info13010027} {Automatic {Curation} of
  {Court} {Documents}: {Anonymizing} {Personal} {Data}}.
\newblock \emph{Information}, 13(1):27.

\bibitem[{Glaser et~al.(2021)Glaser, Schamberger, and
  Matthes}]{glaser_anonymization_2021}
Ingo Glaser, Tom Schamberger, and Florian Matthes. 2021.
\newblock \href {https://doi.org/10.1145/3462757.3466087} {Anonymization of
  german legal court rulings}.
\newblock In \emph{Proceedings of the {Eighteenth} {International} {Conference}
  on {Artificial} {Intelligence} and {Law}}, pages 205--209, São Paulo Brazil.
  ACM.

\bibitem[{Henderson et~al.(2022)Henderson, Krass, Zheng, Guha, Manning,
  Jurafsky, and Ho}]{henderson_pile_2022}
Peter Henderson, Mark~S. Krass, Lucia Zheng, Neel Guha, Christopher~D. Manning,
  Dan Jurafsky, and Daniel~E. Ho. 2022.
\newblock \href {http://arxiv.org/abs/2207.00220} {Pile of {Law}: {Learning}
  {Responsible} {Data} {Filtering} from the {Law} and a {256GB} {Open}-{Source}
  {Legal} {Dataset}}.
\newblock ArXiv:2207.00220 [cs].

\bibitem[{Honnibal et~al.(2020)Honnibal, Montani, Van~Landeghem, and
  Boyd}]{honnibal_spacy_2020}
Matthew Honnibal, Ines Montani, Sofie Van~Landeghem, and Adriane Boyd. 2020.
\newblock \href {https://doi.org/10.5281/zenodo.1212303} {{spaCy}:
  {Industrial}-strength {Natural} {Language} {Processing} in {Python}}.

\bibitem[{Hua et~al.(2022)Hua, Zhang, Chen, Li, and
  Weber}]{hua_legalrelectra_2022}
Wenyue Hua, Yuchen Zhang, Zhe Chen, Josie Li, and Melanie Weber. 2022.
\newblock \href {https://doi.org/10.48550/arXiv.2212.08204} {{LegalRelectra}:
  {Mixed}-domain {Language} {Modeling} for {Long}-range {Legal} {Text}
  {Comprehension}}.
\newblock ArXiv:2212.08204 [cs].

\bibitem[{Kitaev et~al.(2020)Kitaev, Kaiser, and
  Levskaya}]{kitaev_reformer_2020}
Nikita Kitaev, Łukasz Kaiser, and Anselm Levskaya. 2020.
\newblock \href {http://arxiv.org/abs/2001.04451} {Reformer: {The} {Efficient}
  {Transformer}}.
\newblock \emph{arXiv:2001.04451 [cs, stat]}.
\newblock ArXiv: 2001.04451.

\bibitem[{Leitner et~al.(2020)Leitner, Rehm, and
  Moreno-Schneider}]{leitner_dataset_2020}
Elena Leitner, Georg Rehm, and Julián Moreno-Schneider. 2020.
\newblock \href {http://arxiv.org/abs/2003.13016} {A {Dataset} of {German}
  {Legal} {Documents} for {Named} {Entity} {Recognition}}.
\newblock \emph{arXiv:2003.13016 [cs]}.
\newblock ArXiv: 2003.13016.

\bibitem[{Luz~de Araujo et~al.(2018)Luz~de Araujo, de~Campos, de~Oliveira,
  Stauffer, Couto, and Bermejo}]{luz_de_araujo_lener-br_2018}
Pedro~Henrique Luz~de Araujo, Teófilo~E. de~Campos, Renato R.~R. de~Oliveira,
  Matheus Stauffer, Samuel Couto, and Paulo Bermejo. 2018.
\newblock {LeNER}-{Br}: {A} {Dataset} for {Named} {Entity} {Recognition} in
  {Brazilian} {Legal} {Text}.
\newblock In \emph{Computational {Processing} of the {Portuguese} {Language}},
  Lecture {Notes} in {Computer} {Science}, pages 313--323, Cham. Springer
  International Publishing.

\bibitem[{Mao and Liu(2019)}]{mao_hadoken_2019}
Jihang Mao and Wanli Liu. 2019.
\newblock \href {https://ceur-ws.org/Vol-2421/MEDDOCAN\_paper\_11.pdf}
  {Hadoken: a {BERT}-{CRF} {Model} for {Medical} {Document} {Anonymization}}.
\newblock In \emph{Proceedings of the {Iberian} {Languages} {Evaluation}
  {Forum} co-located with 35th {Conference} of the {Spanish} {Society} for
  {Natural} {Language} {Processing}, {IberLEF}@{SEPLN} 2019, {Bilbao}, {Spain},
  {September} 24th, 2019}, volume 2421 of \emph{{CEUR} {Workshop}
  {Proceedings}}, pages 720--726. CEUR-WS.org.

\bibitem[{Niklaus et~al.(2021)Niklaus, Chalkidis, and
  Stürmer}]{niklaus_swiss-judgment-prediction_2021}
Joel Niklaus, Ilias Chalkidis, and Matthias Stürmer. 2021.
\newblock \href {https://aclanthology.org/2021.nllp-1.3}
  {Swiss-{Judgment}-{Prediction}: {A} {Multilingual} {Legal} {Judgment}
  {Prediction} {Benchmark}}.
\newblock In \emph{Proceedings of the {Natural} {Legal} {Language} {Processing}
  {Workshop} 2021}, pages 19--35, Punta Cana, Dominican Republic. Association
  for Computational Linguistics.

\bibitem[{Niklaus and Giofré(2022)}]{niklaus_budgetlongformer_2022}
Joel Niklaus and Daniele Giofré. 2022.
\newblock \href {https://doi.org/10.48550/arXiv.2211.17135}
  {{BudgetLongformer}: {Can} we {Cheaply} {Pretrain} a {SotA} {Legal}
  {Language} {Model} {From} {Scratch}?}
\newblock ArXiv:2211.17135 [cs].

\bibitem[{Niklaus et~al.(2023{\natexlab{a}})Niklaus, Matoshi, Rani, Galassi,
  Stürmer, and Chalkidis}]{niklaus2023lextreme}
Joel Niklaus, Veton Matoshi, Pooja Rani, Andrea Galassi, Matthias Stürmer, and
  Ilias Chalkidis. 2023{\natexlab{a}}.
\newblock \href {http://arxiv.org/abs/2301.13126} {Lextreme: A multi-lingual
  and multi-task benchmark for the legal domain}.

\bibitem[{Niklaus et~al.(2023{\natexlab{b}})Niklaus, Matoshi, Stürmer,
  Chalkidis, and Ho}]{niklaus_multilegalpile_2023}
Joel Niklaus, Veton Matoshi, Matthias Stürmer, Ilias Chalkidis, and Daniel~E.
  Ho. 2023{\natexlab{b}}.
\newblock \href {http://arxiv.org/abs/2306.02069} {{MultiLegalPile}: {A}
  {689GB} {Multilingual} {Legal} {Corpus}}.
\newblock ArXiv:2306.02069 [cs].

\bibitem[{Niklaus et~al.(2022)Niklaus, Stürmer, and
  Chalkidis}]{niklaus_empirical_2022}
Joel Niklaus, Matthias Stürmer, and Ilias Chalkidis. 2022.
\newblock \href {https://aclanthology.org/2022.aacl-main.3} {An {Empirical}
  {Study} on {Cross}-{X} {Transfer} for {Legal} {Judgment} {Prediction}}.
\newblock In \emph{Proceedings of the 2nd {Conference} of the {Asia}-{Pacific}
  {Chapter} of the {Association} for {Computational} {Linguistics} and the 12th
  {International} {Joint} {Conference} on {Natural} {Language} {Processing}
  ({Volume} 1: {Long} {Papers})}, pages 32--46, Online only. Association for
  Computational Linguistics.

\bibitem[{Nyffenegger et~al.(2023)Nyffenegger, Stürmer, and
  Niklaus}]{nyffenegger_anonymity_2023}
Alex Nyffenegger, Matthias Stürmer, and Joel Niklaus. 2023.
\newblock \href {http://arxiv.org/abs/2308.11103} {Anonymity at {Risk}?
  {Assessing} {Re}-{Identification} {Capabilities} of {Large} {Language}
  {Models}}.
\newblock ArXiv:2308.11103 [cs].

\bibitem[{Ouyang et~al.(2022)Ouyang, Wu, Jiang, Almeida, Wainwright, Mishkin,
  Zhang, Agarwal, Slama, Ray, Schulman, Hilton, Kelton, Miller, Simens, Askell,
  Welinder, Christiano, Leike, and Lowe}]{instructgpt}
Long Ouyang, Jeff Wu, Xu~Jiang, Diogo Almeida, Carroll~L. Wainwright, Pamela
  Mishkin, Chong Zhang, Sandhini Agarwal, Katarina Slama, Alex Ray, John
  Schulman, Jacob Hilton, Fraser Kelton, Luke Miller, Maddie Simens, Amanda
  Askell, Peter Welinder, Paul Christiano, Jan Leike, and Ryan Lowe. 2022.
\newblock \href {https://doi.org/10.48550/ARXIV.2203.02155} {Training language
  models to follow instructions with human feedback}.

\bibitem[{Pais et~al.(2021)Pais, Mitrofan, Gasan, Coneschi, and
  Ianov}]{pais_named_2021}
Vasile Pais, Maria Mitrofan, Carol~Luca Gasan, Vlad Coneschi, and Alexandru
  Ianov. 2021.
\newblock \href {https://doi.org/10.18653/v1/2021.nllp-1.2} {Named {Entity}
  {Recognition} in the {Romanian} {Legal} {Domain}}.
\newblock In \emph{Proceedings of the {Natural} {Legal} {Language} {Processing}
  {Workshop} 2021}, pages 9--18, Punta Cana, Dominican Republic. Association
  for Computational Linguistics.

\bibitem[{Pfeiffer et~al.(2021)Pfeiffer, Vuli{\'c}, Gurevych, and
  Ruder}]{pfeiffer-etal-2021-unks}
Jonas Pfeiffer, Ivan Vuli{\'c}, Iryna Gurevych, and Sebastian Ruder. 2021.
\newblock \href {https://doi.org/10.18653/v1/2021.emnlp-main.800} {{UNK}s
  everywhere: {A}dapting multilingual language models to new scripts}.
\newblock In \emph{Proceedings of the 2021 Conference on Empirical Methods in
  Natural Language Processing}, pages 10186--10203, Online and Punta Cana,
  Dominican Republic. Association for Computational Linguistics.

\bibitem[{Raffel et~al.(2020)Raffel, Shazeer, Roberts, Lee, Narang, Matena,
  Zhou, Li, and Liu}]{raffel_exploring_2020}
Colin Raffel, Noam Shazeer, Adam Roberts, Katherine Lee, Sharan Narang, Michael
  Matena, Yanqi Zhou, Wei Li, and Peter~J. Liu. 2020.
\newblock \href {http://jmlr.org/papers/v21/20-074.html} {Exploring the
  {Limits} of {Transfer} {Learning} with a {Unified} {Text}-to-{Text}
  {Transformer}}.
\newblock \emph{Journal of Machine Learning Research}, 21(140):1--67.

\bibitem[{Rasiah et~al.(2023)Rasiah, Stern, Matoshi, Stürmer, Chalkidis, Ho,
  and Niklaus}]{rasiah_scale_2023}
Vishvaksenan Rasiah, Ronja Stern, Veton Matoshi, Matthias Stürmer, Ilias
  Chalkidis, Daniel~E. Ho, and Joel Niklaus. 2023.
\newblock \href {https://doi.org/10.48550/arXiv.2306.09237} {{SCALE}: {Scaling}
  up the {Complexity} for {Advanced} {Language} {Model} {Evaluation}}.
\newblock ArXiv:2306.09237 [cs].

\bibitem[{Ruch et~al.(2000)Ruch, Baud, Rassinoux, Bouillon, and
  Robert}]{ruch_medical_2000}
P.~Ruch, R.~H. Baud, A.~M. Rassinoux, P.~Bouillon, and G.~Robert. 2000.
\newblock Medical document anonymization with a semantic lexicon.
\newblock \emph{Proceedings. AMIA Symposium}, pages 729--733.

\bibitem[{Wei et~al.(2021)Wei, Bosma, Zhao, Guu, Yu, Lester, Du, Dai, and
  Le}]{wei-etal-2022}
Jason Wei, Maarten Bosma, Vincent~Y. Zhao, Kelvin Guu, Adams~Wei Yu, Brian
  Lester, Nan Du, Andrew~M. Dai, and Quoc~V. Le. 2021.
\newblock \href {http://arxiv.org/abs/2109.01652} {Finetuned language models
  are zero-shot learners}.
\newblock \emph{CoRR}, abs/2109.01652.

\bibitem[{Wettig et~al.(2023)Wettig, Gao, Zhong, and Chen}]{wettig2022should}
Alexander Wettig, Tianyu Gao, Zexuan Zhong, and Danqi Chen. 2023.
\newblock \href {https://aclanthology.org/2023.eacl-main.217} {Should you mask
  15{\%} in masked language modeling?}
\newblock In \emph{Proceedings of the 17th Conference of the European Chapter
  of the Association for Computational Linguistics}, pages 2985--3000,
  Dubrovnik, Croatia. Association for Computational Linguistics.

\bibitem[{Zheng et~al.(2021)Zheng, Guha, Anderson, Henderson, and
  Ho}]{zheng_when_2021}
Lucia Zheng, Neel Guha, Brandon~R. Anderson, Peter Henderson, and Daniel~E. Ho.
  2021.
\newblock \href {http://arxiv.org/abs/2104.08671} {When {Does} {Pretraining}
  {Help}? {Assessing} {Self}-{Supervised} {Learning} for {Law} and the
  {CaseHOLD} {Dataset}}.
\newblock \emph{arXiv:2104.08671 [cs]}.
\newblock ArXiv: 2104.08671 version: 3.

\end{thebibliography}

\appendix

\end{document}